\documentclass[preprint,12pt]{elsarticle}

\usepackage[margin=1in]{geometry}   
\usepackage{cite}
\usepackage{amsmath,amssymb,amsfonts}
\usepackage{algorithm}
\usepackage[noend]{algpseudocode}
\usepackage{graphicx}
\usepackage{caption}
\usepackage[table]{xcolor}
\usepackage{textcomp}
\usepackage{booktabs}
\usepackage{tabularx}
\usepackage{array}
\usepackage{hyperref}                
\hypersetup{colorlinks=true,linkcolor=black,citecolor=black,urlcolor=black}

\journal{Nuclear Physics B}

\begin{document}
% \maketitle

\begin{frontmatter}
\title{Artifact Removal and Image Restoration in AFM:\\
A Structured Mask-Guided Directional Inpainting Approach}

\author[label1]{Juntao Zhang}
\author[label2]{Angona Biswas}
\author[label2]{Jaydeep Rade}
\author[label1]{Charchit Shukla}
\author[label1]{Juan Ren\thanks{Corresponding author: \texttt{juanren@iastate.edu}}}
\author[label2]{Anwesha Sarkar}
\author[label1]{Adarsh Krishnamurthy}
\author[label1]{Aditya Balu}
\affiliation[label1]{organization={Department of Mechanical Engineering},%Department and Organization
            addressline={Iowa State University}, 
            city={Ames},
            postcode={50011}, 
            state={IA},
            country={USA}}
\affiliation[label2]{organization={Department of Electrical and Computer Engineering},%Department and Organization
            addressline={Iowa State University}, 
            city={Ames},
            postcode={50011}, 
            state={IA},
            country={USA}}
\begin{abstract}
Atomic Force Microscopy (AFM) enables high-resolution surface imaging at the nanoscale, yet the output is often degraded by artifacts introduced by environmental noise, scanning imperfections, and tip-sample interactions. To address this challenge, a lightweight and fully automated framework for artifact detection and restoration in AFM image analysis is presented. The pipeline begins with a classification model that determines whether an AFM image contains artifacts. If necessary, a lightweight semantic segmentation network, custom-designed and trained on AFM data, is applied to generate precise artifact masks. These masks are adaptively expanded based on their structural orientation and then inpainted using a directional neighbor-based interpolation strategy to preserve 3D surface continuity. A localized Gaussian smoothing operation is then applied for seamless restoration. The system is integrated into a user-friendly GUI that supports real-time parameter adjustments and batch processing. Experimental results demonstrate the effective artifact removal while preserving nanoscale structural details, providing a robust, geometry-aware solution for high-fidelity AFM data interpretation.
\end{abstract}

%% Keywords

\end{frontmatter}

\section{Introduction}

Atomic Force Microscopy (AFM) is a key imaging technique for characterizing surface topography at the nanoscale \citep{kaupp2006atomic, farokh2023imaging}. By scanning a cantilever probe across a sample surface and recording vertical deflections, AFM provides three-dimensional height maps with nanometer-level resolution. This capability makes it indispensable in materials science, nanotechnology, and biophysics \citep{ando2022high, puthukodan2023purification}. However, the quality and interpretability of AFM images are often compromised by artifacts originating from environmental disturbances, scanner nonlinearities, and probe-sample interactions \citep{joo2025atomic, bellotti2022afm, kocur2023correction, liu2024revolutionizing, paruchuri2024machine}. These artifacts, which include scan line distortions, streaks, pits, and sample-induced anomalies, can significantly hinder accurate surface analysis.

Existing AFM artifact removal methods typically fall into two categories: manual correction through software tools, such as Bruker's Nanoscope Analysis, and algorithmic restoration using polynomial fitting, median filtering, or kriging interpolation \citep{jahnavi2024comparative,arildsen2015reconstruction, chen2011destripe,shukla2023quantification}. While these approaches offer partial solutions, they either require expert supervision or lack generalizability across artifact types. Furthermore, most conventional inpainting techniques operate on 2D image intensity, ignoring the intrinsic 3D height structure of AFM data. Deep learning-based image restoration techniques, including CNN- and GAN-based models, have also been explored in both general image inpainting and limited AFM-related contexts\citep{zhai2023comprehensive,gao2020generative}.

In AFM-specific contexts, a growing body of research has demonstrated the potential of AI-driven methods. Early efforts introduced AI-guided AFM measurements of live cells \citep{rade2021ai, masud2024machine}, followed by the application of deep learning for live cell shape detection and automated AFM navigation \citep{rade2022deep}. More recently, CNN-based models have been developed for AFM imaging defect detection and classification \citep{zhang2024afm, 11175706}, while large language model (LLM)-based decision support has been proposed to assist AFM defect classification workflows \citep{biswas2025conversational}. Together, these studies highlight the increasing integration of machine learning into AFM imaging, yet most remain task-specific, computationally demanding, or limited in their ability to generalize across artifact types.

To overcome these limitations, this study proposes a lightweight, geometry-aware, and fully automated pipeline that unifies classification, segmentation, and inpainting-based restoration of AFM data. The entire process begins with raw SPM files, which are converted into 2D grayscale PNG representations. A convolutional neural network (CNN) based classification model is then used to automatically determine whether an AFM image contains artifacts. Specifically, images are classified into three categories: \textit{good} (artifact-free), \textit{not tracking} (due to scanner instability or loss of feedback), and \textit{tip contamination} (resulting from probe defects that cause duplication or smearing). This multi-class classification strategy allows for early screening of AFM datasets. Images flagged as clean are directly exported, while those flagged as corrupted proceed to the restoration module.

For the restoration stage, a lightweight semantic segmentation model, specifically tailored and trained on AFM datasets, is employed to generate binary masks identifying artifact regions. These masks are further processed through adaptive expansion based on aspect ratio and spatial orientation, ensuring comprehensive artifact coverage. A directional neighbor-based interpolation technique is applied to fill in masked regions using structurally relevant neighboring height values, preserving topographical continuity. Finally, a localized Gaussian smoothing operation refines the transition between inpainted and unmasked regions, improving visual coherence and data reliability.

To enhance usability and support flexible parameter tuning, the proposed pipeline is encapsulated in a user-friendly graphical user interface (GUI). Built using the Tkinter framework \citep{moore2018python}, the GUI allows users to import raw SPM files, visualize 2D/3D AFM images, and interactively configure processing parameters. The interface includes real-time previews of image flattening results, segmentation masks, and inpainted surfaces through a six-panel visualization layout. Users can choose between different flattening strategies (line-by-line or mask-aware), adjust segmentation thresholds, filter artifact regions based on area and aspect ratio, and apply directional inpainting using the Telea method \citep{telea2004image}. The GUI not only provides convenient visualization but also ensures reproducibility of outputs through consistent export of height vectors.

\begin{figure*}[!t]
    \centering
    \includegraphics[width=0.95\textwidth]{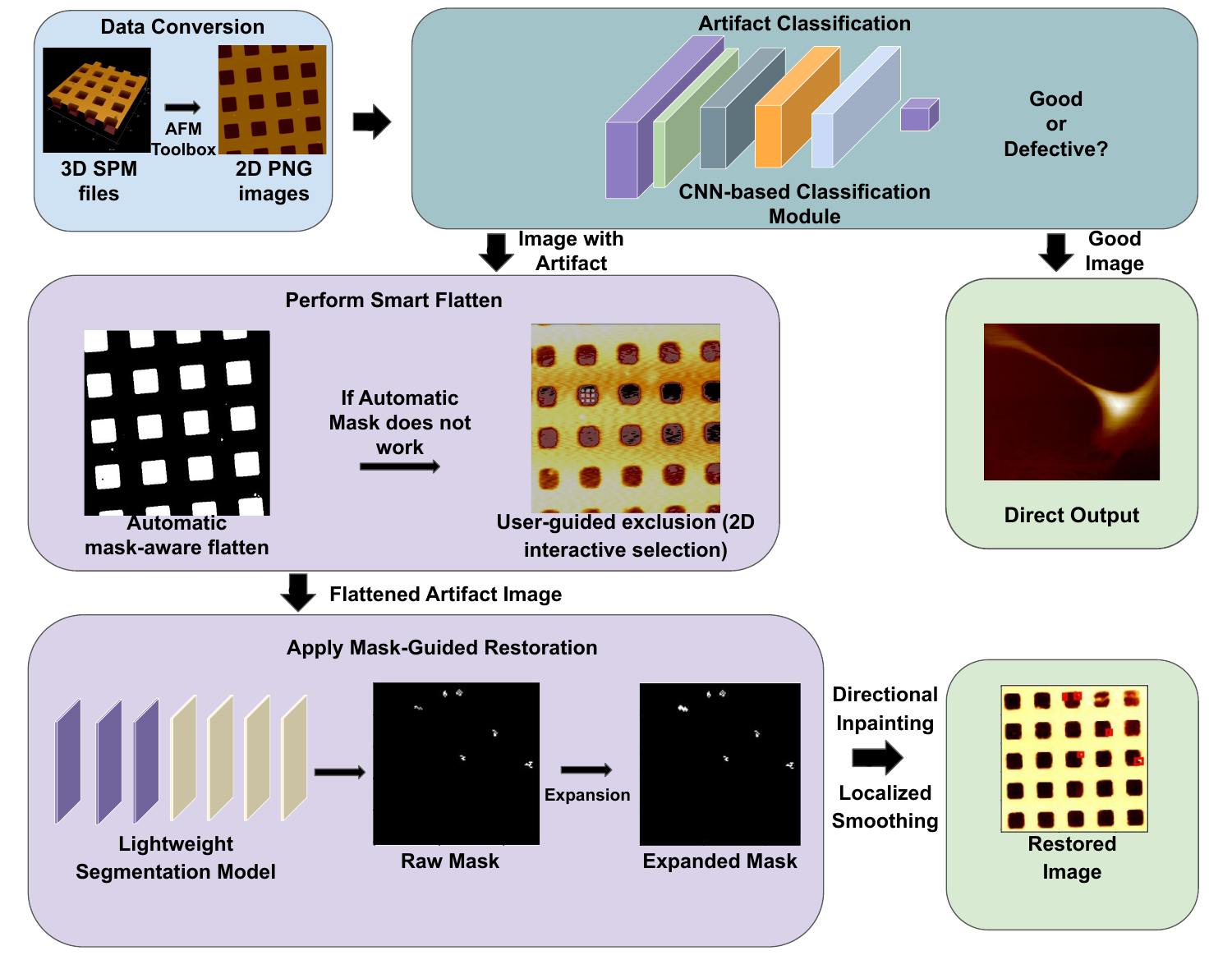}
    \caption{
    Overview of the proposed AFM image restoration framework. 
    Raw SPM data are converted to 2D height maps for CNN-based artifact classification. 
    Defective images undergo Smart Flatten, which supports automatic mask-aware 
    or user-guided interactive exclusion, followed by Mask-Guided Restoration 
    via segmentation-based mask expansion, directional inpainting, and localized smoothing.
    }
    \label{fig:workflow}
\end{figure*}

As illustrated in Figure~\ref{fig:workflow}, the proposed AFM image processing framework integrates image classification, semantic segmentation, artifact-aware restoration, and interactive visualization in a unified pipeline. The classification model enables the early-stage filtering of artifact-free images, thereby avoiding unnecessary computation. The segmentation module ensures fast and precise artifact localization. The mask-guided directional inpainting maintains nanoscale topographical continuity, while the GUI empowers users with real-time control and intuitive feedback. Together, these modules form a complete solution for high-throughput, artifact-aware AFM data analysis.

\section{Methodology}

The proposed AFM image analysis framework integrates automated classification, lightweight semantic segmentation, smart flattening, mask-guided directional inpainting, and an interactive graphical user interface (GUI). The workflow begins with raw SPM files and ends with either fully restored AFM height maps or annotated artifact masks.
\subsection{Workflow Overview}

As illustrated in Figure~\ref{fig:workflow}, the proposed AFM image restoration framework integrates both \textit{automated} and 
\textit{user-guided} modes to balance high-throughput processing with expert-driven precision. The pipeline begins with raw SPM data conversion and CNN-based artifact classification, 
followed by a hybrid Smart Flatten module that can operate in either automatic 
mask-aware or interactive exclusion mode. This dual design allows users to manually 
refine flattening regions when automated masks are insufficient, ensuring robustness 
across diverse AFM datasets and artifact types. After flattening, the system applies 
\textbf{Mask-Guided Restoration} through segmentation-based mask expansion, 
directional inpainting, and localized smoothing to reconstruct high-fidelity AFM surfaces.

The complete workflow consists of five major components. First, proprietary SPM files are parsed using a custom SPMDataExtractor class that interfaces with Bruker’s DataSourceDLL.dll via Python \texttt{ctypes}, extracting scan size, resolution, and height matrices with consistent unit scaling. 

Next, a CNN-based artifact classification module distinguishes between good and defective AFM scans. Images identified as artifact-free are directly exported, while defective scans proceed to the restoration pipeline. 

Flattening is then performed using the proposed Smart Flatten approach, which supports direction-aware, mask-aware baseline removal as well as optional user-guided interactive exclusion. This step removes tilt and scan-line distortions while preserving nanoscale surface morphology. 

Finally, mask-guided restoration is applied by generating and expanding artifact masks using a lightweight segmentation model, followed by directional inpainting and localized smoothing to reconstruct surface continuity. All modules are integrated within a Tkinter-based graphical user interface that consolidates file I/O, parameter control, visualization, and result export, enabling real-time interaction with both automatic and user-guided modes.

\subsection{Artifact Classification Module}

\begin{figure*}[!b]
    \centering
    \includegraphics[width=1.00\textwidth]{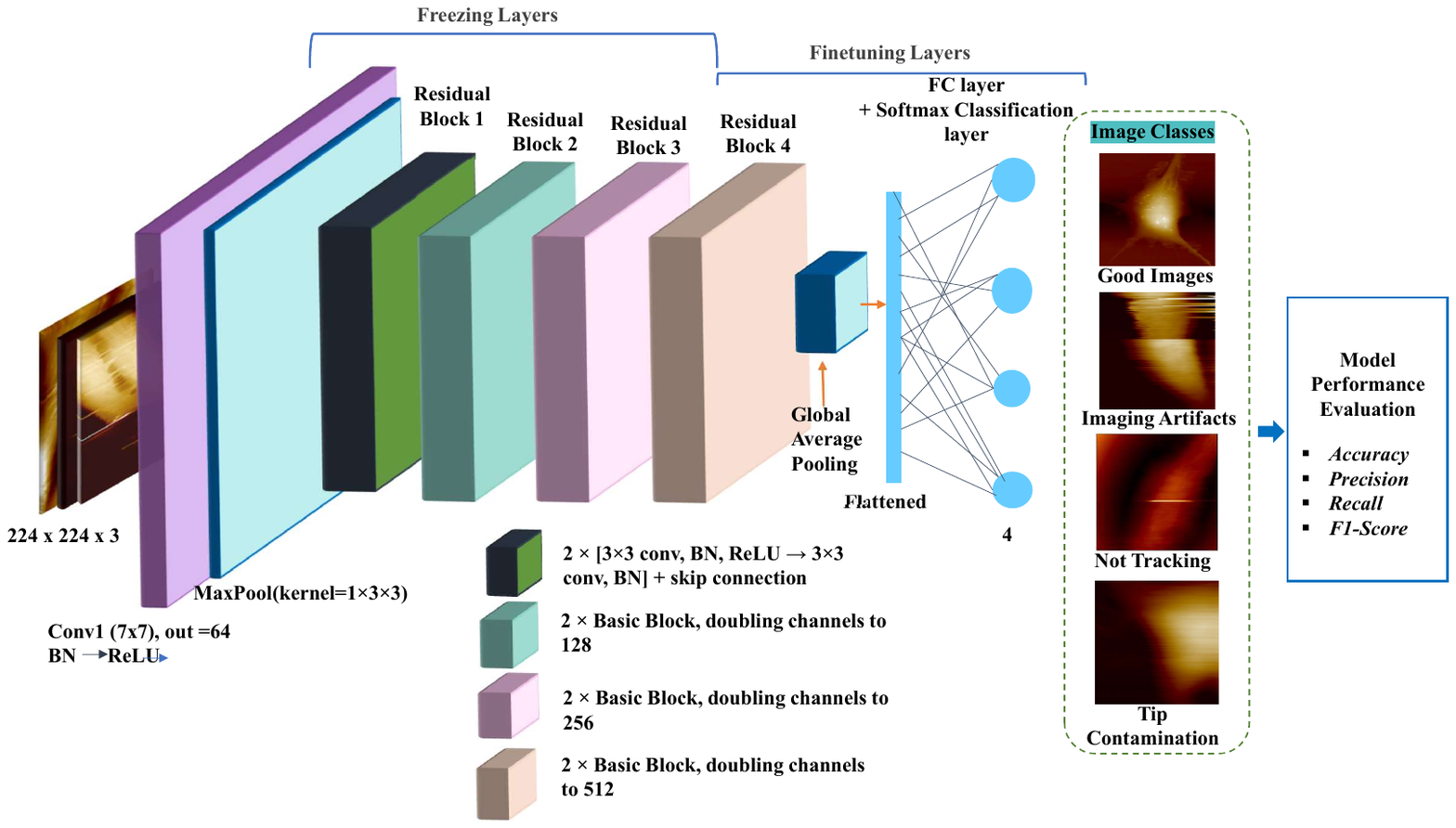}
    \caption{CNN-based classification module for AFM artifact detection. 
    The model is based on ResNet-18 pretrained on ImageNet, with only the final residual block and fully connected (FC) layer fine-tuned for AFM data. 
    Four classes are distinguished: Good, Imaging Artifacts, Not Tracking, and Tip Contamination. 
    Evaluation is based on accuracy, precision, recall, and F1-score. 
    The classification stage acts as a decision gate: \textit{Good} images are directly exported, while corrupted ones are passed to the subsequent restoration pipeline.}
    \label{fig:classification_model}
\end{figure*}

AFM images often contain various defects such as not-tracking stripes, tip contamination, or other imaging artifacts. To automatically identify whether an input AFM image is suitable for further analysis, a residual CNN-based classification module was implemented (Figure~\ref{fig:classification_model}). The classifier takes PNG images of size $224\times224\times3$ converted from raw SPM height data and outputs one of four labels: \textit{Good}, \textit{Not Tracking}, \textit{Tip Contamination}, or \textit{Imaging Artifacts}. 

The backbone network is based on ResNet-18 pretrained on ImageNet. To adapt the model for AFM image classification, the final fully connected (FC) layer was replaced with a four-class classifier, and only the last residual block and FC layer were fine-tuned, while earlier convolutional layers were frozen. This transfer learning strategy reduces overfitting and accelerates convergence, given the limited size of AFM datasets. Training employed Focal Loss to address class imbalance, with optimization performed using stochastic gradient descent (SGD, learning rate $10^{-4}$, momentum $0.9$) and adaptive scheduling via ReduceLROnPlateau. Metrics such as accuracy, precision, recall, and F1-score were used for evaluation. 

Importantly, the classification stage serves as a decision point in the overall workflow. Images predicted as \textit{Good} are directly exported without further modification, while images classified as corrupted (\textit{Not Tracking}, \textit{Tip Contamination}, or \textit{Imaging Artifacts}) are routed to the subsequent segmentation, flattening, and restoration pipeline.

\begin{algorithm}[H]
\caption{AFM Defect Classification using ResNet-18}
\label{alg:classification}
\begin{algorithmic}[1]
\State \textbf{Input:} AFM SPM height data $\rightarrow$ PNG images $(224 \times 224 \times 3)$
\State \textbf{Output:} Class label $y \in \{\text{Good}, \text{Not Tracking}, \text{Tip Contamination}, \text{Imaging Artifacts}\}$

\State \textbf{Step 1: Data Preparation} 
\State \quad Convert raw AFM scans to image patches 
\State \quad Normalize: $z = (x - \mu)/\sigma$ (dataset statistics) 
\State \quad Apply augmentation: rotation, flip, zoom, color jitter 
\State \quad Split dataset: Train (75\%), Validation (15\%), Test (10\%) 

\State \textbf{Step 2: Model Initialization} 
\State \quad Load ResNet-18 pretrained on ImageNet 
\State \quad Replace final FC layer with 4-class classifier 
\State \quad Freeze $\{\text{Conv1, Blocks1--3}\}$, unfreeze $\{\text{Block4, FC}\}$ 

\State \textbf{Step 3: Training} 
\For{each epoch}
    \For{each batch $(X,y)$}
        \State $\hat{y} = \text{ResNet-18}(X)$ 
        \State $\mathcal{L} = \text{FocalLoss}(\hat{y}, y, w_c)$ 
        \State Update $\theta$ using SGD ($lr=10^{-4}$, momentum=0.9) 
        \State Adjust $lr$ via ReduceLROnPlateau 
    \EndFor
\EndFor

\State \textbf{Step 4: Fine-Tuning Strategy} 
\State \quad Progressive unfreezing: $\{\text{FC}\} \rightarrow \{\text{Block4}\}$ 
\State \quad Select best configuration via validation accuracy 

\State \textbf{Step 5: Evaluation} 
\State \quad Compute Accuracy, Precision, Recall, F1 
\State \quad Deploy optimal model $M^*$ on unseen test set 
\end{algorithmic}
\end{algorithm}
\subsection{Mask Generation Module}

\begin{figure*}[!b]
    \centering
    \includegraphics[width=1.05\textwidth]{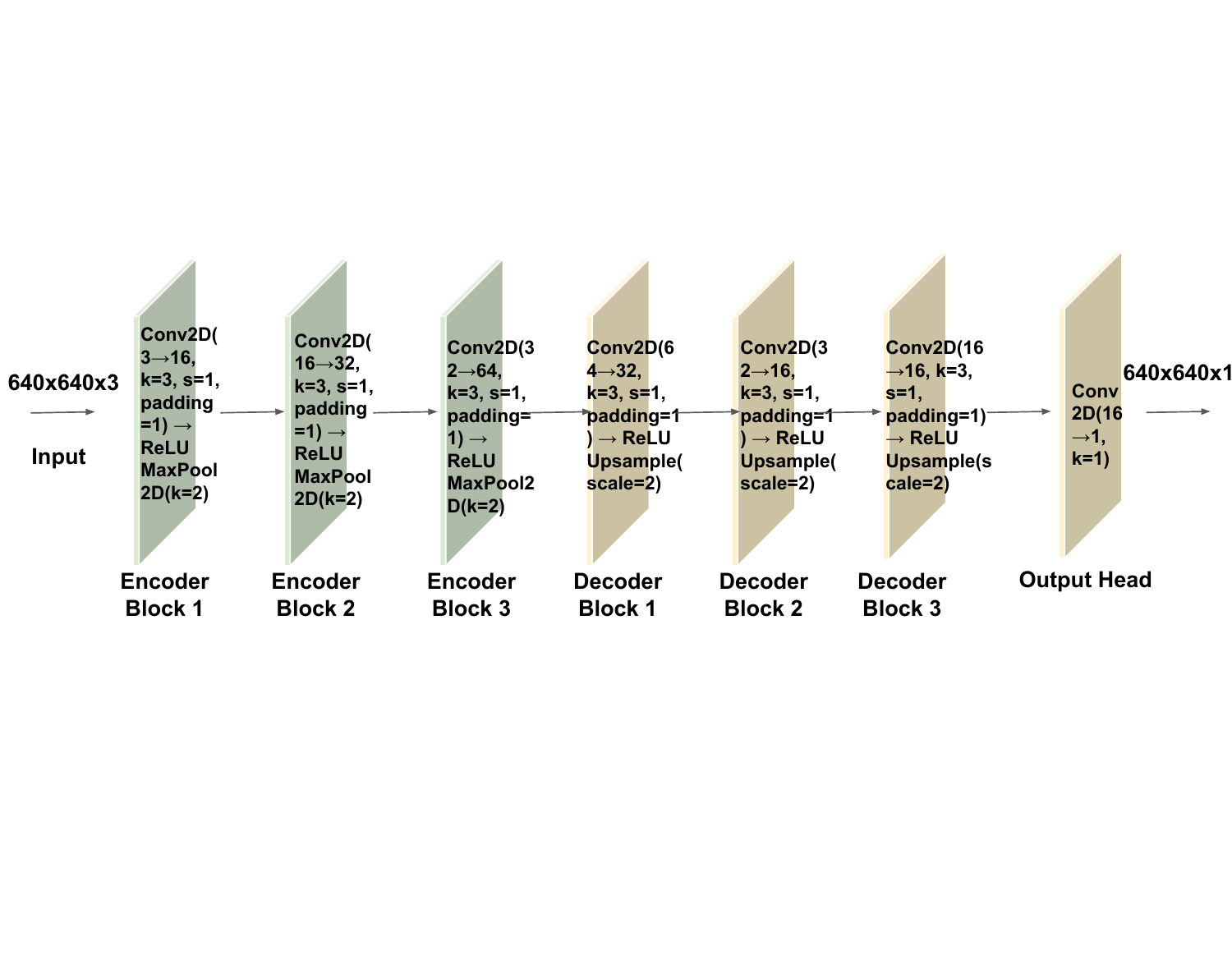}
    \vspace{-10.0em}
    \caption{Lightweight semantic segmentation model for AFM artifact mask generation.}
    \label{fig:segmentation_model}
\end{figure*}

For images classified as corrupted, artifact localization is achieved through a custom, lightweight semantic segmentation model \citep{wu2022uiu, deng2022elu}, as shown in Figure~\ref{fig:segmentation_model}. The model outputs a normalized response map \(P(i,j)\in[0,1]\) for each pixel, indicating the likelihood of artifact presence. Unless otherwise specified, the segmentation and restoration modules use a consistent set of tunable parameters. Table~\ref{tab:gui_parameters} summarizes the key steps and parameters of the mask generation module. Specifically, the raw artifact probability map is converted into a binary mask using a threshold parameter $\texttt{thr} \in [0,1]$. Small spurious detections and large regular structures are removed using area constraints $\texttt{min\_pix}$ and $\texttt{max\_pix}$. For each connected component, an aspect ratio threshold $\texttt{ar}$ is used to distinguish elongated stripe-like artifacts from compact blob-like regions.

During stripe expansion, similarity-based growth is controlled by an intensity tolerance factor $\texttt{k}$, a local neighborhood size $\texttt{region}$, and a gradient limit $\texttt{grad}$, ensuring that expansion remains confined to regions with similar height statistics. For spike suppression, isolated outliers are detected within a sliding window of size $\texttt{win}$ using a deviation criterion $|h - \mathrm{median}| > \lambda \sigma$, where $\lambda = \texttt{lam}$. In the restoration stage, the inpainting radius $\texttt{radius}$ determines the spatial extent of directional and Telea-based interpolation.
\begin{table}[htbp]
\centering
\caption{Mask generation key steps and parameters.}
\label{tab:gui_parameters}
\begin{tabularx}{\textwidth}{>{\hsize=0.8\hsize}X
                                >{\hsize=1.2\hsize}X
                                >{\hsize=1.0\hsize}X
                                >{\hsize=1.0\hsize}X}

\toprule
\textbf{Step} & \textbf{Purpose} & \textbf{Key Parameters} & \textbf{Outcome} \\
\midrule
1. Lightweight segmentation pre-screen & CNN predicts suspicious/normal pixels from the height map & \texttt{thr} & Raw mask \\
2. Despike merge & Flag isolated spikes where $|h - \mathrm{median}| > \lambda \sigma$ & \texttt{win}, \texttt{lam} & Raw mask $\cup$ \texttt{mask\_spike} \\
3. Area filter & Remove small noise and large regular structures & \texttt{min\_pix}, \texttt{max\_pix} & Area-filtered mask \\
4. Stripe seed extraction & Compute aspect ratio per component & \texttt{ar} & Stripe seed / blob \\
5. Stripe similarity growth & Expand stripe seeds using $\mu \pm k\sigma$ and gradient limit & \texttt{k}, \texttt{region}, \texttt{grad} & Expanded stripe mask \\
6. Final mask assembly & Combine expanded stripes and blobs & --- & Final mask \\
7. Class-specific action & Apply inpainting or overlay based on class type & \texttt{radius} & Modified image / overlay \\
\bottomrule
\end{tabularx}
\end{table}
\noindent A binary raw mask \(M_{\mathrm{raw}}(i,j)\) is generated by thresholding the network response map
using a threshold parameter \(\tau_{\mathrm{thr}}\):

\begin{equation}
M_{\mathrm{raw}}(i,j) =
\begin{cases}
1, & \text{if } P(i,j) \ge \tau_{\mathrm{thr}}, \\
0, & \text{otherwise}.
\end{cases}
\end{equation}

\noindent Then, to remove spurious detections, connected components in \(M_{\mathrm{raw}}\) are filtered by area:
\begin{equation}
    M_{\mathrm{filt}} = \{ \mathcal{C} \in M_{\mathrm{raw}} \ |\ \mathrm{min\_pix} \le A(\mathcal{C}) \le \mathrm{max\_pix} \},
\end{equation}
where \(A(\mathcal{C})\) is the pixel area of connected component \(\mathcal{C}\), and \(\mathrm{min\_pix}\), \(\mathrm{max\_pix}\) are adjustable.

\noindent For stripe and blob separation, each connected component \(\mathcal{C}\) in \(M_{\mathrm{filt}}\), the aspect ratio is computed as:
\begin{equation}
    \mathrm{AR}(\mathcal{C}) = \frac{\text{width}(\mathcal{C})}{\text{height}(\mathcal{C})}.
\end{equation}
Components are classified as:
\[
\text{stripe if } \mathrm{AR}(\mathcal{C}) \ge \mathrm{ar\_thr}, \quad \text{blob otherwise}.
\]
The threshold \(\mathrm{ar\_thr}\) is adjustable as well.

\noindent Stripes are extended using a similarity-based expansion strategy
(\texttt{expand\_by\_similarity}) with parameters \(k\), \(\mathrm{grad\_th}\), and \texttt{region}.
Specifically, for each candidate pixel \(q\) adjacent to the current stripe region \(\mathcal{S}\),
the following conditions are evaluated:

\begin{equation}
    |I(q) - \mu_{\mathcal{S}}| \le k \cdot \sigma_{\mathcal{S}} \quad \text{and} \quad \nabla I(q) \le \mathrm{grad\_th}.
\end{equation}

\noindent The final expanded mask is obtained as:
\begin{equation}
    M_{\mathrm{exp}} = \mathcal{S}_{\mathrm{exp}} \ \cup \ \mathcal{B}.
\end{equation}

\

\subsection{Smart Flatten Module}

The flattening stage removes low-frequency background trends and tilt from AFM topography images. 
The \textbf{Smart Flatten} approach incorporates:  
(1) \textbf{Mask-aware fitting} to bypass defect regions,  
(2) \textbf{Orientation flexibility} allowing row-wise or column-wise flattening,  
(3) Support for low-order polynomial fitting,  
(4) \textbf{User-guided exclusion} through an interactive 2D interface.

In the automatic mode, Smart Flatten operates in a fully mask-aware and direction-adaptive manner. 
For each scan line, polynomial fitting is performed using only unmasked pixels, and the fitting direction (row or column) is determined based on the dominant slope direction.

To further improve robustness, particularly in heterogeneous samples where automated 
masks may be insufficient, an interactive mode is provided via the GUI. Users can 
manually select regions directly on the 2D AFM image, and those selected pixels are 
excluded from baseline fitting. This user-guided exclusion ensures that step edges, 
unusual structures, or mislabeled artifacts do not distort the flattening result. 
By combining automatic mask-aware fitting with optional expert-guided region selection, 
the Smart Flatten module achieves both high automation and fine-grained control.

For flexibility, all flattening parameters (direction, polynomial order, mask-aware enable/disable, 
and user-guided exclusion) are adjustable.

\begin{figure*}[h]
    \centering
    \includegraphics[width=0.95\textwidth]{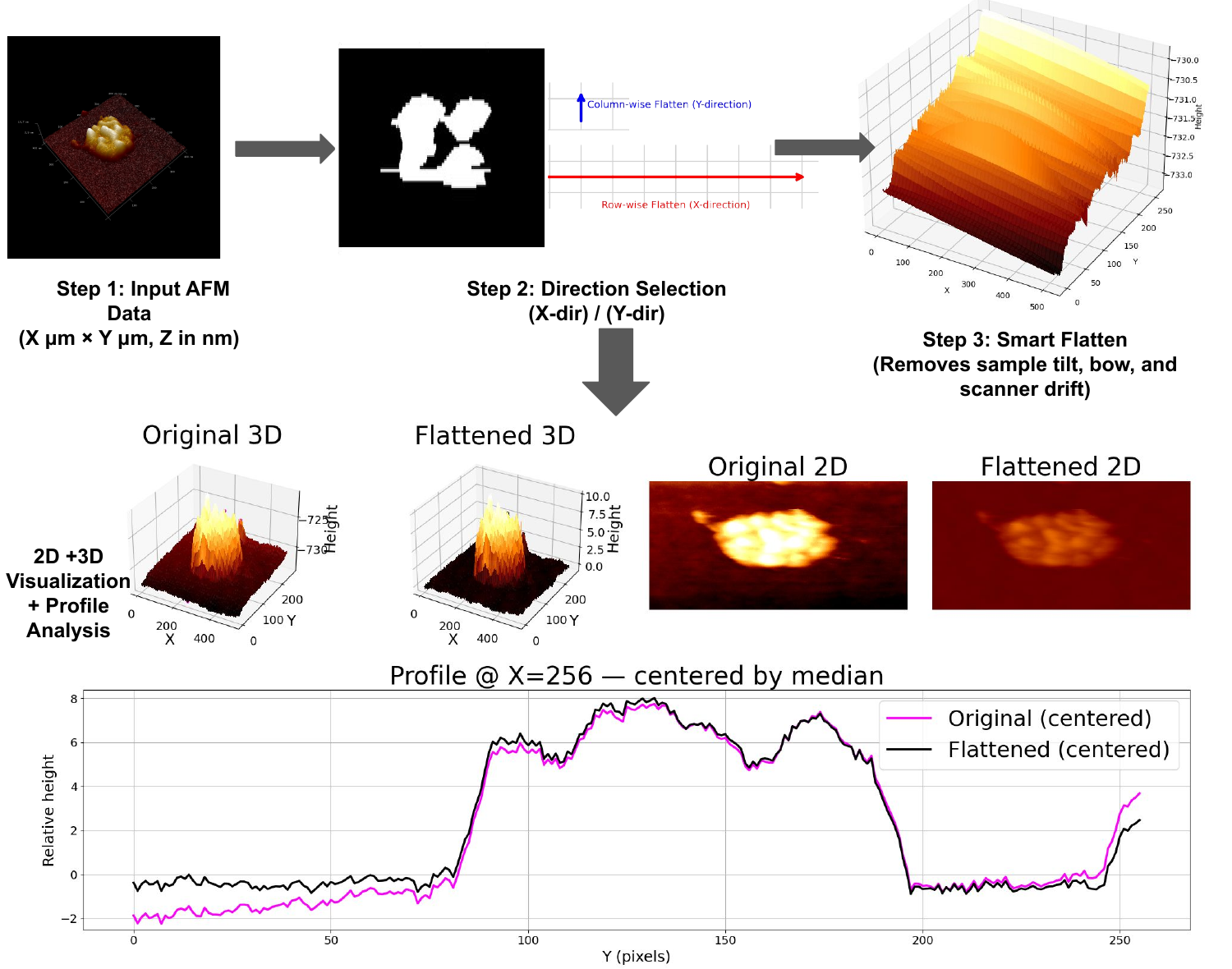}
    \caption{Schematic of the Smart Flatten process.}
    \label{fig:smart_flatten}
\end{figure*}

\subsection{Mask-Guided Restoration}

Following the flattening stage, the artifact mask is further expanded and used to guide the restoration process. To ensure comprehensive coverage of scan-line artifacts, elongated stripe components are first expanded according to their aspect ratio, with full-row expansion applied to horizontal stripes and full-column expansion applied to vertical stripes. In addition, a similarity-based growth strategy is employed to extend stripe regions based on local intensity statistics and gradient constraints, allowing the mask to adaptively capture artifact regions with similar height characteristics.

Restoration is then performed using a directional neighbor-based inpainting strategy, in which masked pixels are filled using the nearest valid neighbors along the dominant scan direction. For extended stripe-like artifacts, OpenCV’s Telea inpainting method is applied, with the inpainting radius adjustable through the GUI to accommodate artifacts of different widths and intensities. Finally, localized Gaussian smoothing is applied exclusively within masked regions to suppress residual discontinuities and ensure smooth transitions between restored and unmodified areas. Together, these steps reconstruct surface continuity while preserving genuine nanoscale features outside artifact regions.

\subsection{GUI Integration}

\begin{figure*}[!t]
    \centering
    \includegraphics[width=1.05\textwidth]{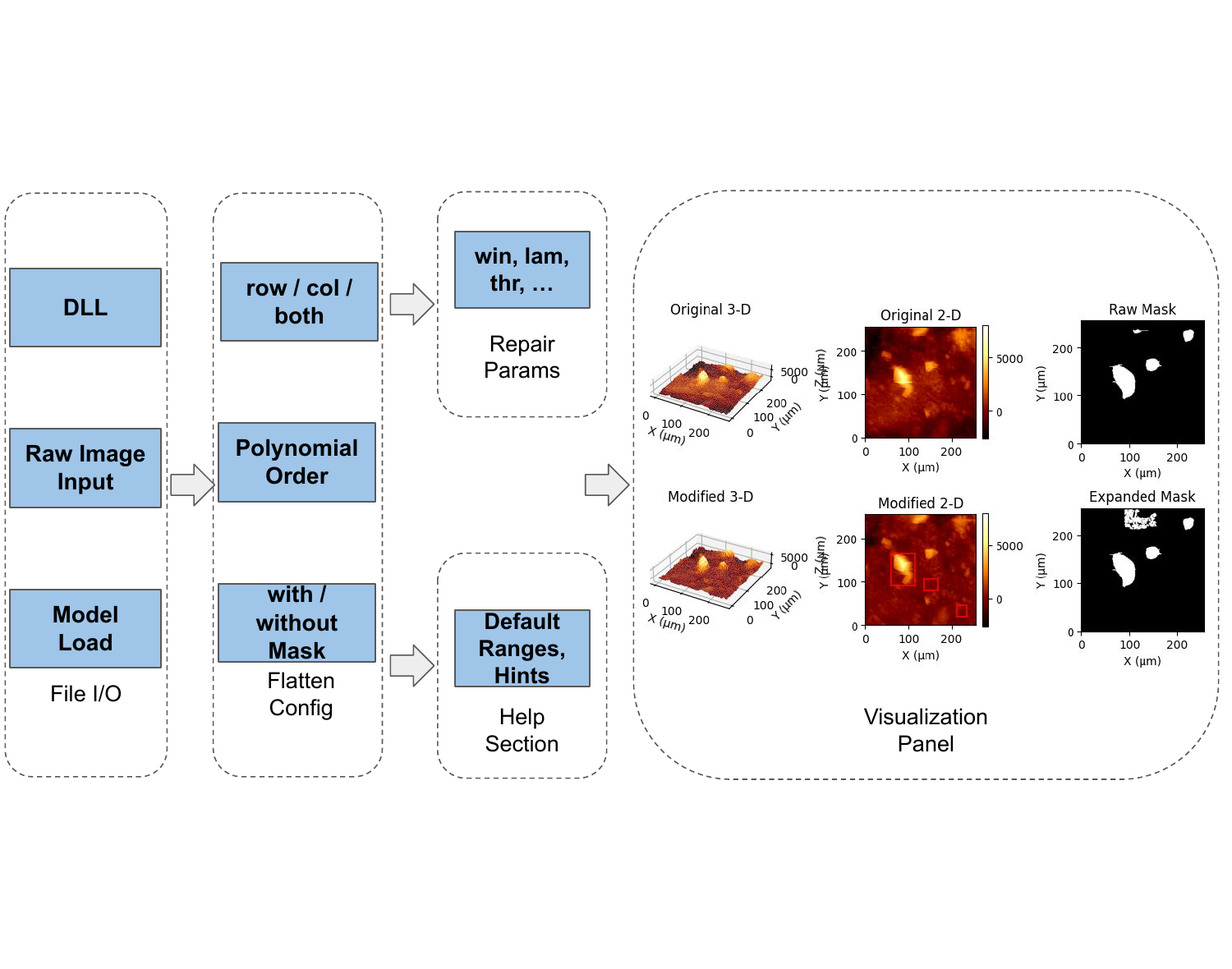}
    \caption{Schematic of the AFM Repair GUI workflow. The interface integrates file I/O for raw SPM data, DLL access, and model loading; configurable flattening options (row/column/bidirectional fitting, polynomial order, mask-aware mode); centralized parameter adjustment with default ranges and tuning hints; and a visualization panel displaying original and modified AFM data (2D/3D views, raw artifact mask, and expanded mask).}
    \label{fig:gui_flow}
\end{figure*}

A Tkinter-based graphical user interface (GUI) was developed to integrate all modules of the proposed AFM artifact detection and restoration pipeline into a unified workflow (Figure~\ref{fig:gui_flow}). The interface streamlines the process from file loading to result export, providing immediate visual feedback throughout each processing stage.

The GUI supports comprehensive file and model management, including loading raw SPM files, specifying the Bruker DLL directory, importing pre-trained model weights, and defining output paths. Users can configure flattening behavior by selecting between standard line-by-line and mask-aware line-by-line fitting, with support for horizontal, vertical, or bidirectional orientations. An interactive user-guided exclusion mode further allows regions to be selected directly on the 2D AFM image and excluded from baseline fitting, preventing structural steps or residual defects from biasing background correction.

Throughout the segmentation, flattening, and restoration stages, a common set of tunable parameters is used and exposed through the GUI. These parameters are consolidated within a centralized \textit{Repair Params} panel, enabling real-time adjustment without modifying source code. The adjustable parameters include the probability threshold \texttt{thr} for mask binarization, the sliding window size \texttt{win} and deviation factor \texttt{lam} for spike detection, the area constraints \texttt{min\_pix} and \texttt{max\_pix} for connected-component filtering, the aspect ratio threshold \texttt{ar} for stripe identification, the similarity tolerance \texttt{k}, local neighborhood size \texttt{region}, and gradient limit \texttt{grad} for stripe expansion, as well as the inpainting radius \texttt{radius} controlling the spatial extent of restoration.

An integrated help section provides descriptions, default values, and recommended ranges for each parameter, along with practical tuning guidelines for different artifact scenarios. The interface further features a six-panel visualization layout that simultaneously displays the original 2D and 3D height maps, raw and expanded artifact masks, the modified 2D height map, and the restored 3D surface. This layout facilitates direct before–after comparison and qualitative assessment. Finally, restored AFM height matrices can be exported in \texttt{.txt} format for downstream analysis or archival purposes.
\subsubsection{Parameter initialization and selection.}
All adjustable parameters in the proposed pipeline are initialized using physics-informed heuristics and empirical observations from AFM imaging practice. Specifically, thresholds related to height variation, gradient magnitude, and connected-component area are selected based on typical noise levels, scan resolution, and artifact morphology observed in representative AFM datasets. 

Initial values are determined using a small validation subset, where parameters are tuned to achieve stable artifact localization and background correction without over-smoothing or structural distortion. Importantly, the selected defaults are not optimized for a single dataset but are chosen to generalize across different samples and scan conditions.

To accommodate dataset heterogeneity and varying artifact characteristics, all parameters remain user-adjustable through the GUI. This design enables expert users to refine parameter settings when automated defaults are insufficient, while preserving robust performance under default configurations.

\section{Results and Discussion}

Both qualitative and quantitative evaluations were conducted focusing on the proposed AFM image classification, artifact detection, and restoration pipeline. The results demonstrate the capability of the integrated framework to automatically identify defective scans, accurately localize artifact regions, and restore surface continuity without compromising nanoscale details.

\subsection{Qualitative Evaluation}
This subsection focuses on a qualitative assessment of the proposed approach, highlighting visual performance and interpretability at different processing stages. Representative AFM examples are used to illustrate classification outcomes and to demonstrate how artifact localization, flattening, and restoration affect surface morphology in both 2D and 3D views.

\subsubsection{ResNet-driven Transfer Learning Framework for Image Defect Classification}

The first stage of the proposed pipeline utilizes a lightweight residual CNN classifier based on ResNet-18 \citep{7780459}, adapted to AFM imaging through transfer learning. 
The model was fine-tuned from ImageNet-pretrained weights using a layer-wise strategy, where early convolutional layers were frozen while the final residual block and fully connected (FC) layers were updated. 
Input data consisted of $224\times224\times3$ PNG images converted from raw SPM height maps. 
The classifier distinguishes among multiple AFM defect types, enabling reliable separation between clean scans and common imaging failures.
Clean scans are directly exported, while defective ones are routed to the restoration pipeline for further processing.

\begin{figure*}[!t]
    \centering
    \includegraphics[width=1.1\textwidth]{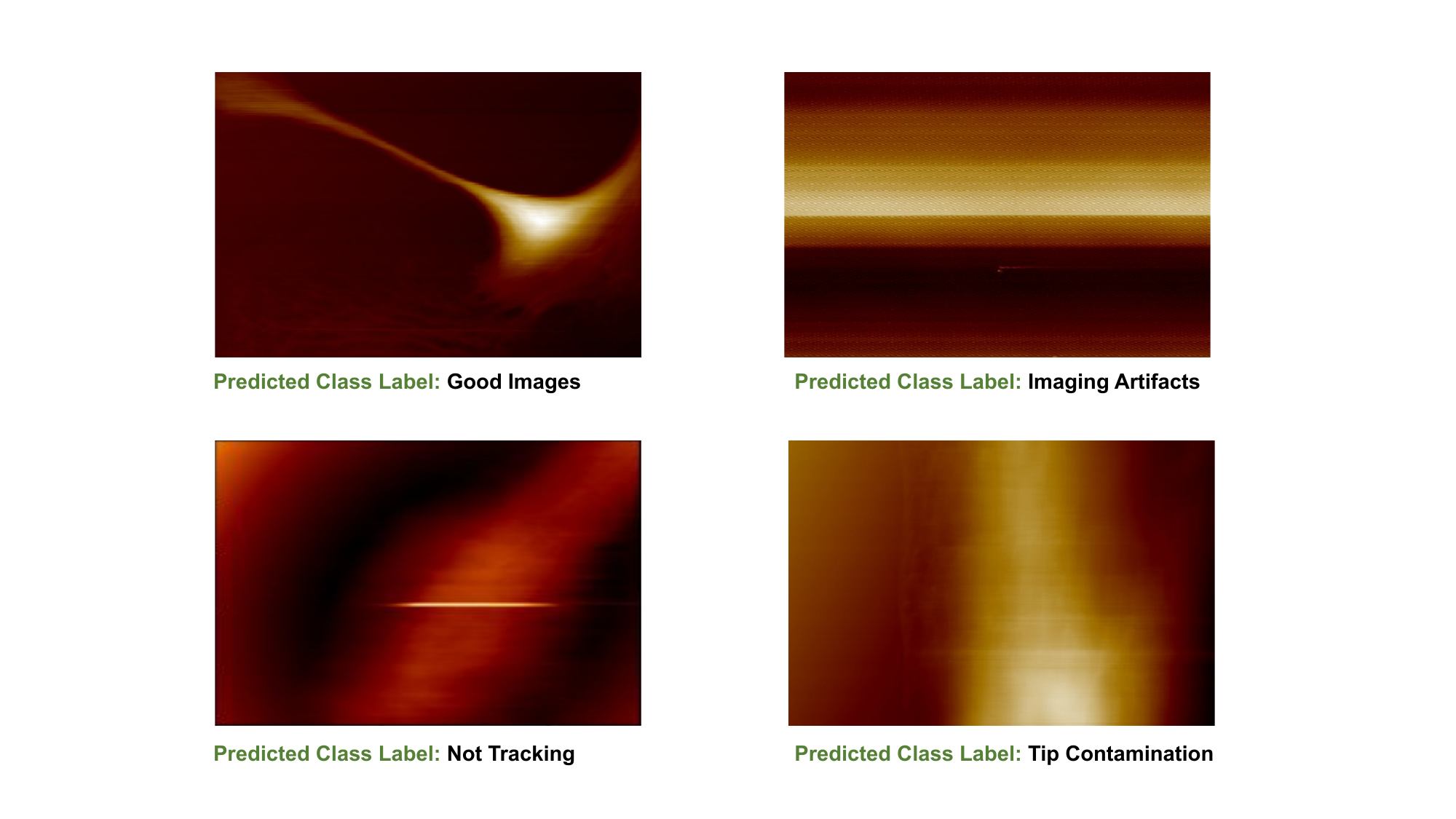}
    \caption{
    Representative classification results for the CNN-based AFM image quality model.
    Four example predictions are shown: \textbf{Good Images}, \textbf{Imaging Artifacts}, 
    \textbf{Not Tracking}, and \textbf{Tip Contamination}.
    Each case illustrates the model’s visual recognition of distinct defect morphologies, 
    accurately distinguishing between clean and corrupted AFM scans.
    }
    \label{fig:classification_examples}
\end{figure*}

\begin{figure*}[!t]
    \centering
    \includegraphics[width=1.1\textwidth]{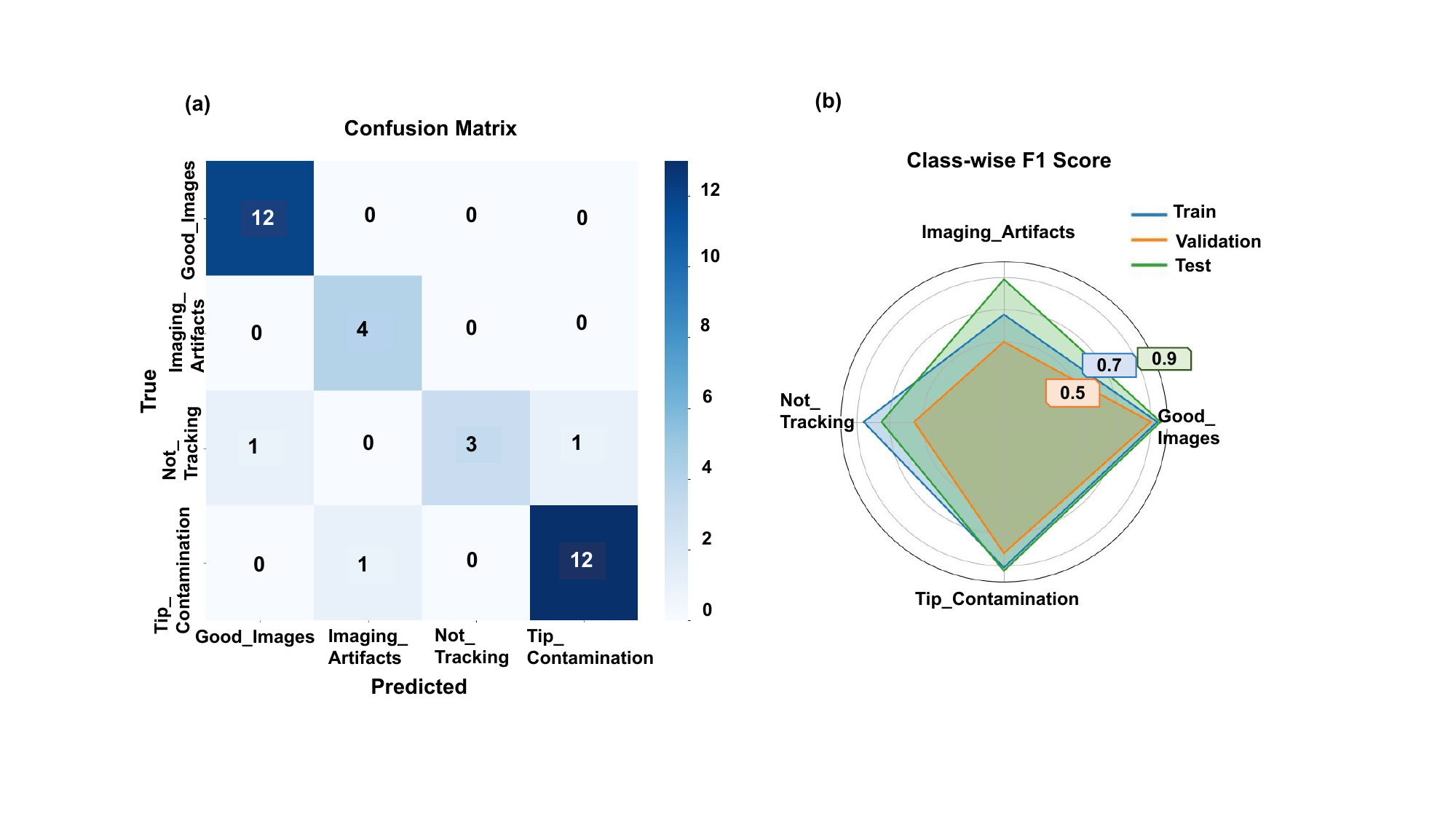}
    \caption{
    Quantitative evaluation of the AFM classification model.
    (a) Confusion matrix showing per-class performance on the test dataset. 
    (b) Class-wise F1-score radar plots comparing training, validation, and test sets. 
    The results demonstrate balanced accuracy and consistent generalization across 
    \textbf{Good Images}, \textbf{Imaging Artifacts}, \textbf{Not Tracking}, 
    and \textbf{Tip Contamination} categories.
    }
    \label{fig:classification_metrics}
\end{figure*}

To improve robustness under limited and imbalanced training data, class-specific data augmentation was applied using geometric transformations and color jitter. 
Training employed Focal Loss \citep{lin2017focal} to mitigate class imbalance, optimized via stochastic gradient descent (SGD, learning rate $10^{-4}$, momentum 0.9) \citep{sclocchi2024different}. 
All datasets were split into train (75\%), validation (15\%), and test (10\%) sets before augmentation to prevent data leakage. 
Normalization was performed independently for AFM data using z-score statistics computed across the dataset.  
The final model achieved a test accuracy of 91.43\%, with perfect recall for \textit{Good Images} (R=1.00) and strong class-wise F1-scores across the remaining defect categories.
Figure~\ref{fig:classification_examples} presents representative classification results for each predicted class, demonstrating that the model correctly identifies distinct artifact types such as line loss and tip duplication.  
Figure~\ref{fig:classification_metrics} summarizes the quantitative evaluation using a confusion matrix and class-wise F1-score radar plots.

The results highlight balanced performance across all four categories, confirming the model’s ability to generalize to unseen AFM defect patterns.

\subsubsection{Artifact Segmentation and Smart Flatten}
\begin{figure*}[!t]
    \centering
    \includegraphics[width=1.0\textwidth]{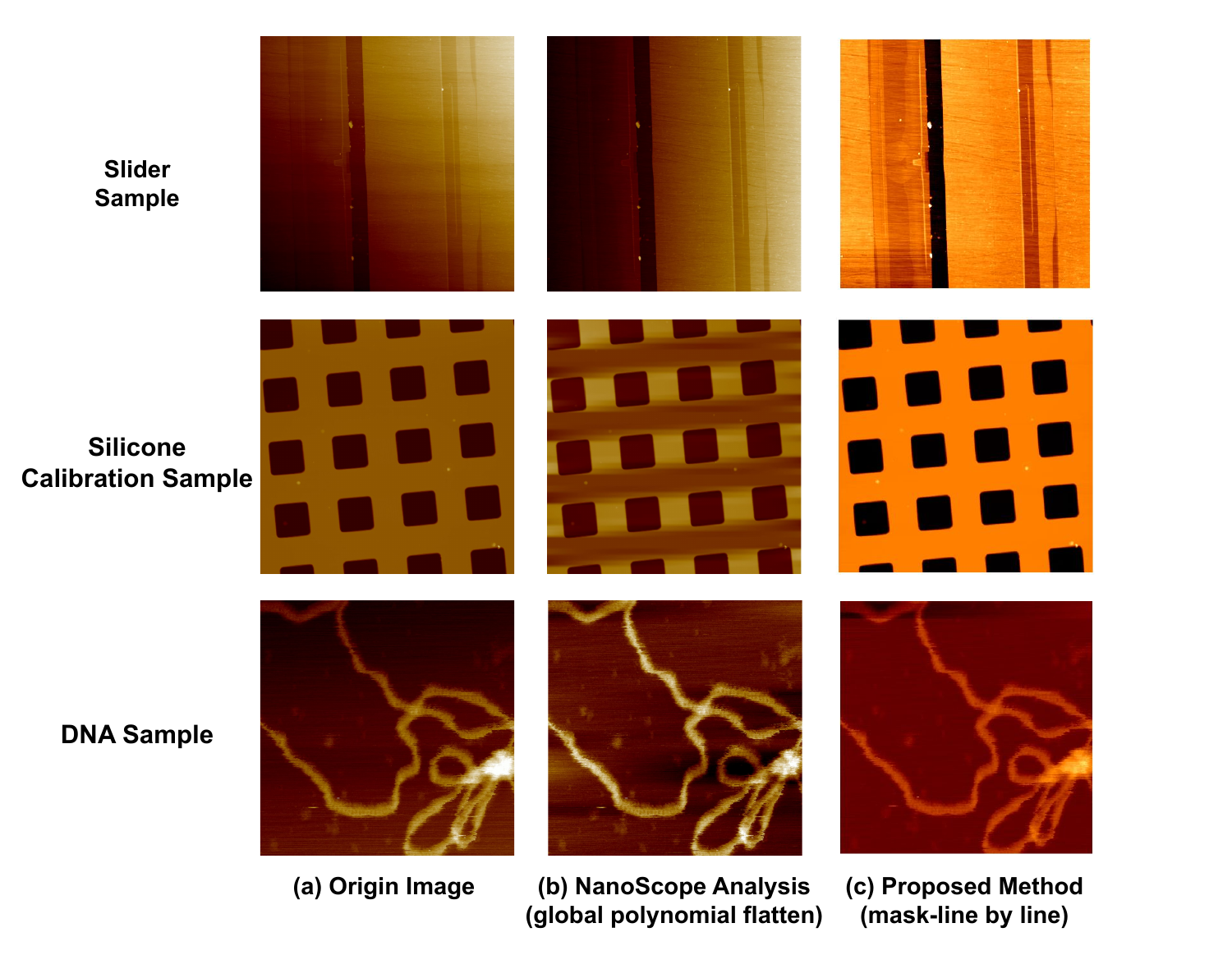}
    \caption{
    Flattening comparison for representative AFM datasets: 
    (Top) slider sample, (Middle) silicone calibration sample, and (Bottom) DNA sample.
    Columns: (a) original image; (b) NanoScope global polynomial flatten; and 
    (c) proposed \textbf{Smart Flatten} (mask-aware, line-by-line, selectable orientation, with optional user-guided exclusion).
    The Smart Flatten method effectively removes tilt and drift while preserving nanoscale features, 
    and the interactive option enables experts to prevent structural steps or residual artifacts from biasing baseline fitting.
    }
    
    \label{fig:flatten_comparison}
\end{figure*}

Defective images are first processed by the lightweight semantic segmentation model to generate binary masks of artifact regions. 
Then the Smart Flatten module removes sample-induced tilt, curvature, and line-wise drift via direction-aware, mask-aware polynomial fitting (horizontal or vertical). In addition to the automatic mask-aware mode, an interactive option enables users to exclude specific regions directly on the 2D AFM image, ensuring that step edges or unrecognized defects do not bias baseline fitting.
Representative flattening comparisons are shown in Fig.~\ref{fig:flatten_comparison}, demonstrating that the proposed method effectively corrects low-frequency background trends while preserving nanoscale features.

\subsubsection{Mask-Guided Restoration}
\begin{figure*}[!t]
    \centering
    \includegraphics[width=1.10\textwidth]{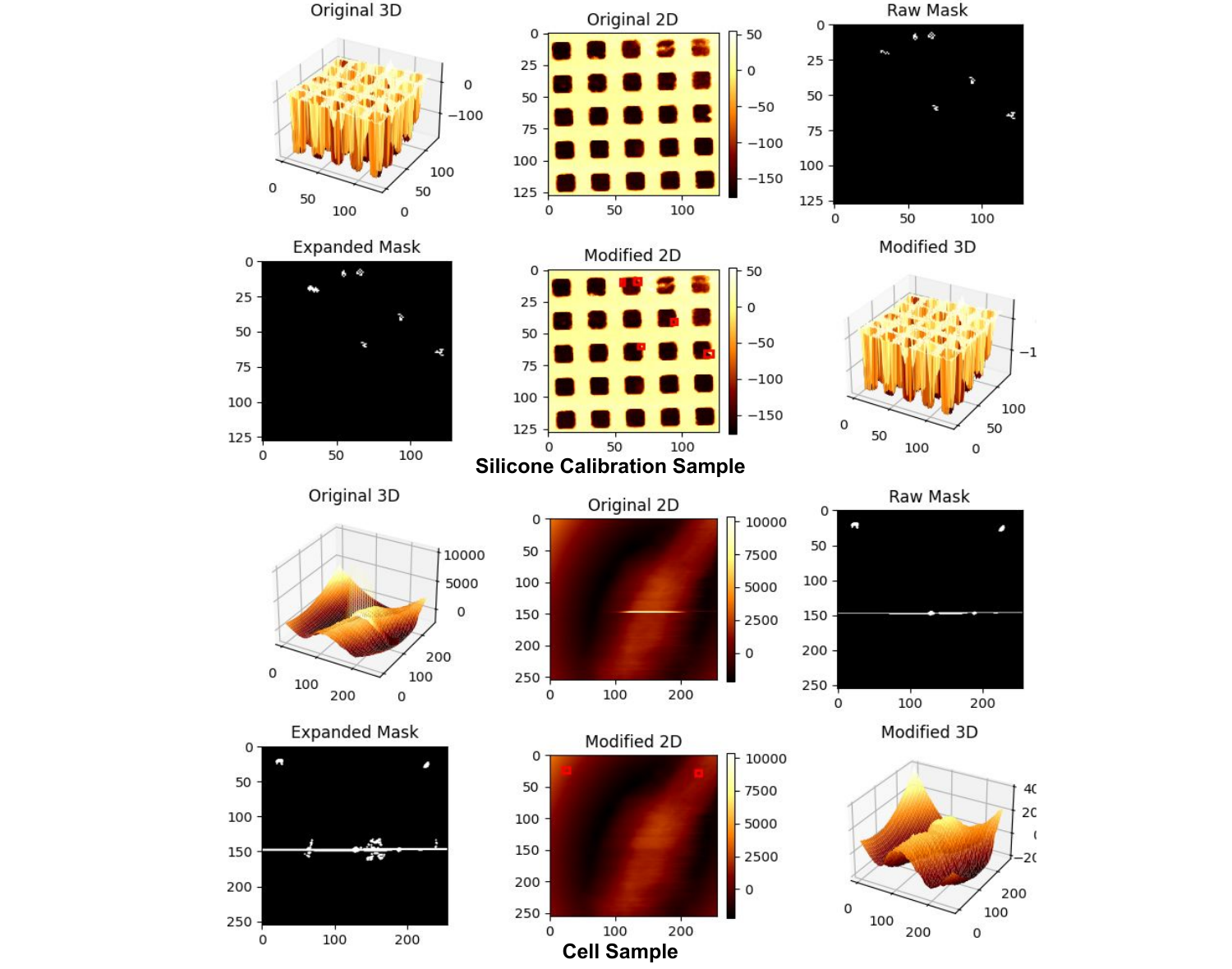}
    \caption{
    Qualitative restoration examples for two representative AFM datasets: 
    (Upper) silicone calibration sample and (Lower) cell sample. 
    For each dataset, the top row shows the original 3D view, original 2D height map, and raw defect mask. 
    The bottom row shows the expanded defect mask (after morphological processing), the restored 2D height map, and the restored 3D view. 
    The process effectively removes scanning artifacts while preserving true surface structures. 
    \textbf{The Smart Flatten stage supports both automatic mask-aware fitting and optional user-guided region exclusion, ensuring reliable background correction across heterogeneous samples.}
    }
    \label{fig:restoration_examples}
\end{figure*}

After flattening, mask-guided directional inpainting and localized smoothing restore surface continuity in artifact regions. 
Visual comparisons in Fig.~\ref{fig:restoration_examples} demonstrate effective removal of elongated scanning streaks and local defects while preserving fine surface structures.

\subsection{Quantitative Evaluation}

\subsubsection{Classification Accuracy Metrics}

The classification performance of the proposed ResNet-18 model demonstrates both accuracy and robustness across all four AFM image categories. 
The network achieved a training accuracy of 89.0\%, validation accuracy of 79.0\%, and a final test accuracy of 91.43\%, indicating effective generalization despite moderate variance between training and validation performance. 
The \textit{Good Images} class achieved the highest recall (R=1.00) and an F1-score of 0.96, confirming the model’s ability to correctly identify clean AFM scans with almost no false negatives. 
The classifier also performed strongly on \textit{Tip Contamination} cases, reaching an F1-score of 0.93, which highlights its robustness in recognizing probe-related artifacts. 
For the \textit{Not Tracking} category, perfect precision (P=1.00) but lower recall (R=0.60) indicated that some streak-like artifacts were visually similar to other defect types, leading to partial overlap in predictions. 
Meanwhile, \textit{Imaging Artifacts} achieved high recall (R=1.00) but slightly reduced precision (P=0.80), suggesting that the model occasionally over-detected ambiguous regions. In 
general, the classifier achieved a Macro-F1 score of 88.25\% and a Weighted-F1 score of 90.91\%, illustrating the effectiveness of transfer learning and focal loss in addressing class imbalance and subtle imaging variations.  

Figure~\ref{fig:classification_metrics} presents the quantitative evaluation of the model through a confusion matrix and a class-wise F1-score radar chart. 
The confusion matrix reveals strong separability across major classes with minimal confusion between similar defect types, while the radar chart confirms consistent performance across training, validation, and test datasets.  
In addition, representative classification outputs for each predicted category are shown in Figure~\ref{fig:classification_examples}, demonstrating that the model accurately captures distinctive morphological patterns for \textit{Good Images}, \textit{Imaging Artifacts}, \textit{Not Tracking}, and \textit{Tip Contamination}.  
Together, these results confirm that the ResNet-18-based classifier provides a reliable front-end filtering mechanism, ensuring that only defective AFM scans are forwarded to the downstream restoration module.

\begin{table}[t!]
    \centering
    \footnotesize
    \caption{Per-class classification performance of the fine-tuned ResNet-18 model with Focal Loss on the AFM defect test dataset.}
    \label{tab:classification_performance}
    \begin{tabular}{l|c|c|c|c}
        \hline
        \textbf{Class} & \textbf{Support} & \textbf{Precision} & \textbf{Recall} & \textbf{F1-Score} \\
        \hline
        Good Images       & 12 & 0.92 & 1.00 & 0.96 \\
        Imaging Artifacts &  4 & 0.80 & 1.00 & 0.89 \\
        Not Tracking      &  5 & 1.00 & 0.60 & 0.75 \\
        Tip Contamination & 14 & 0.93 & 0.93 & 0.93 \\
        \hline
        \textbf{Overall Accuracy} & \multicolumn{4}{c}{91.43\%} \\
        \hline
        \textbf{Macro-F1 Score}   & \multicolumn{4}{c}{88.25\%} \\
        \textbf{Weighted-F1 Score} & \multicolumn{4}{c}{90.91\%} \\
        \hline
    \end{tabular}
\end{table}

\subsubsection{Segmentation Quality}

The performance of artifact localization is evaluated against manual annotations 
using the Intersection-over-Union (IoU) and Dice coefficient. In the test set, the 
lightweight segmentation model achieves an IoU of 0.72 and a Dice coefficient of 0.83, 
indicating a reliable spatial agreement between the predicted artifact masks and the ground truth.

These results demonstrate that the generated masks are sufficiently accurate to 
support the downstream Smart Flatten and mask-guided restoration. Although the model is not 
designed to maximize segmentation accuracy as a standalone task, the overlap achieved 
is adequate to prevent artifact regions from biasing baseline fitting and to ensure 
robust restoration performance.

\subsubsection{Smart Flatten: Ablation and Metrics}
The Smart Flatten module is evaluated on \textit{nonartifact background regions} to ensure that flattening effectively removes tilt and drift without distorting true surface features. To quantitatively assess the effectiveness of Smart Flatten on artifact-free background regions, three complementary metrics are considered: (1) Background variation, measured as the standard deviation of height values ($\sigma_{\mathrm{bg}}$), reflects the residual surface roughness after flattening, with lower values indicating improved background correction. 
(2) Line-wise residual RMSE ($\mathrm{RMSE}_{\mathrm{line}}$) is computed between each scan line and its fitted polynomial baseline, capturing the effectiveness of drift and bow removal along the fast-scan direction. 
(3) Gradient continuity, quantified as the mean absolute gradient magnitude within non-artifact regions ($\Delta|\nabla h|$), indicates the absence of artificial steps or discontinuities introduced during flattening.

As summarized in Table~\ref{tab:flatten_ablation}, Smart Flatten significantly reduces line-wise residuals compared to both global polynomial fitting and single-direction line-by-line flattening, achieving an average $\mathrm{RMSE}_{\mathrm{line}}$ of $49.53 \pm 63.72$~nm vs. $135.67 \pm 188.32$~nm for global polynomial fitting. Although column-only flattening yields the lowest background variation ($134.39 \pm 173.89$~nm), Smart Flatten provides a balanced improvement across all metrics, reducing bias from defect regions and adapting to the dominant slope direction. 

Note that the interactive mode of Smart Flatten allows experts to manually exclude regions that automated masks may miss, ensuring that step edges or anomalous structures do not bias the baseline fitting. This optional user guidance complements the automatic pipeline and further enhances robustness, especially in heterogeneous AFM samples. These results confirm the effectiveness of mask-aware, direction-selectable, and user-guided flattening in enhancing AFM background correction.

\begin{table}[!t]
\centering
\caption{Smart Flatten ablation on background regions (mean $\pm$ std across the test set).}
\label{tab:flatten_ablation}
\begin{tabular}{lccc}
\hline
\textbf{Method} & $\sigma_{\mathrm{bg}}$ (nm) & $\mathrm{RMSE}_{\mathrm{line}}$ (nm) & $\Delta \nabla h$ \\
\hline
Global poly (order=2)      & 348.58 $\pm$ 520.26 & 135.67 $\pm$ 188.32 & 12.03 $\pm$ 18.22 \\
Line-by-line (row only)    & 267.51 $\pm$ 394.12 & 108.56 $\pm$ 151.78 & 13.03 $\pm$ 17.26 \\
Line-by-line (col only)    & 134.39 $\pm$ 173.89 & 63.30 $\pm$ 82.54   & \textbf{9.32 $\pm$ 10.97} \\
\textbf{Smart Flatten (ours)} & 170.09 $\pm$ 249.58 & \textbf{49.53 $\pm$ 63.72}  & 13.80 $\pm$ 19.82 \\
\hline
\end{tabular}
\end{table}

Additionally, we evaluated the \textit{tilt removal ratio} $\rho_{\mathrm{tilt}}$ quantifies the relative reduction 
in least-squares plane slope magnitude between pre- and post-flattening:
\[
\rho_{\mathrm{tilt}} = 1 - \frac{\lVert \nabla \hat{z}_{\mathrm{post}} \rVert_2}{\lVert \nabla \hat{z}_{\mathrm{pre}} \rVert_2},
\]
where $\nabla \hat{z}$ is obtained from a plane fit to the background. 
The proposed method achieves an average $\rho_{\mathrm{tilt}}$ of approximately 85\%, 
substantially outperforming global polynomial and single-direction line flattening (give the values of global polynomial and single direction line flatten) , 
while preserving morphological fidelity.

\subsubsection{Robustness Evaluation under Directional Ripple Noise}
\begin{figure*}[!t]
    \centering
    \includegraphics[width=1.1\textwidth]{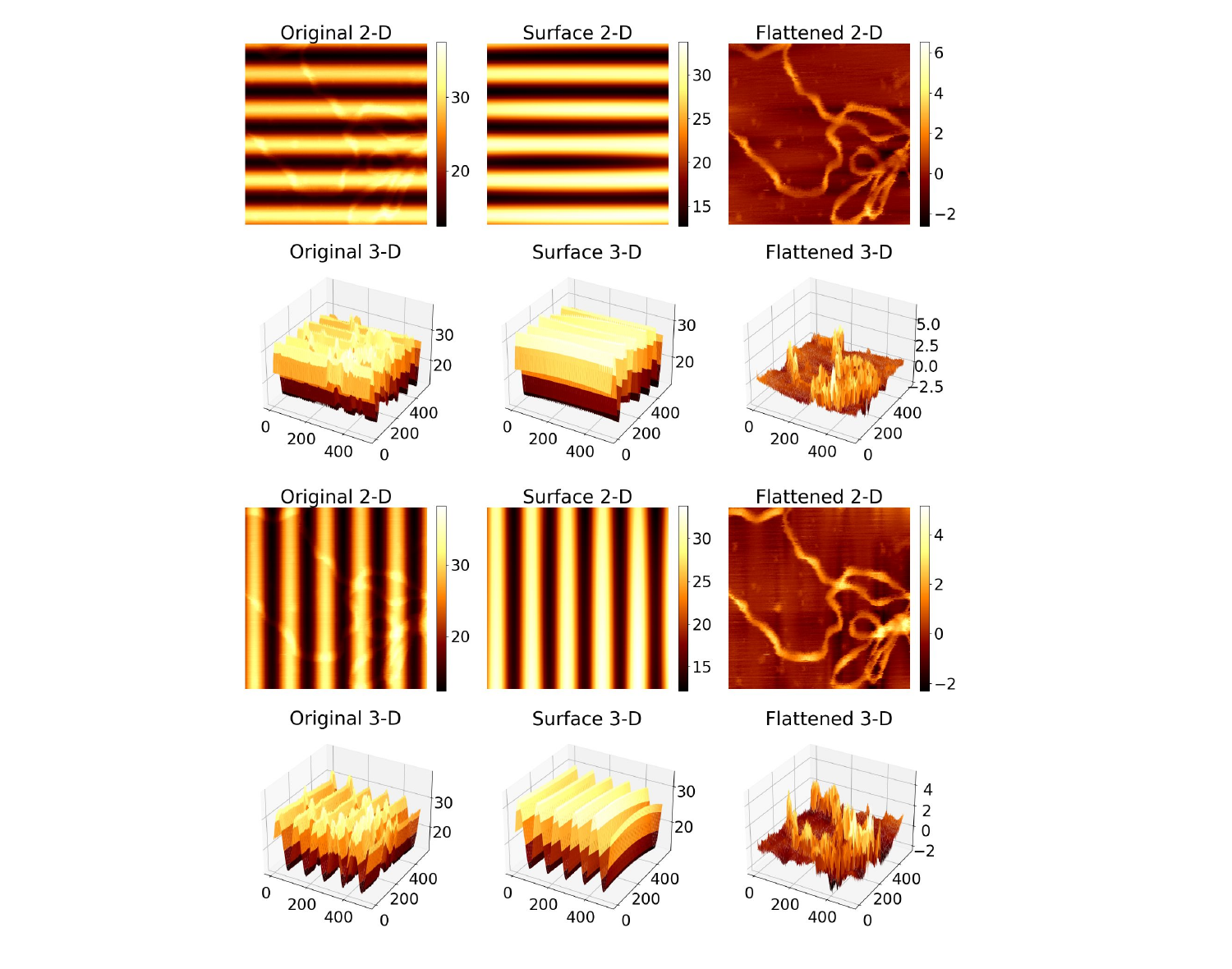}
    \caption{
    Directional robustness evaluation of Smart Flatten under 40\% ripple noise.
    Synthetic sinusoidal interference was applied to a DNA AFM height map 
    with two orientations: (top) horizontal ripple ($\theta = 0^\circ$) 
    and (bottom) vertical ripple ($\theta = 90^\circ$), 
    each at $\alpha = 0.40$, $f = 0.01$~px$^{-1}$.  
    Each triplet shows: (a) original 2D/3D AFM surface, 
    (b) ripple-added image, and (c) flattened result using 
    \textbf{Smart Flatten}.  
    The proposed method effectively removes direction-dependent periodic noise 
    while retaining nanoscale surface features, demonstrating high robustness 
    and adaptability to varying ripple orientations.
    }
    
    \label{fig:ripple40_flatten}
\end{figure*}

To assess the directional robustness of the proposed \textbf{Smart Flatten} algorithm, 
synthetic ripple noise was introduced into clean AFM height maps of the DNA sample.
The perturbation follows a sinusoidal model that emulates periodic scanning distortions:
\[
z_{\mathrm{ripple}}(i,j) = z(i,j) + \alpha |z(i,j)|
\sin\!\big( 2\pi f (i\sin\theta + j\cos\theta) + \phi \big),
\]
where $\alpha$ is the relative amplitude (fraction of local height), 
$f$ the spatial frequency, $\theta$ the ripple orientation, and $\phi$ the phase offset.
This construction maintains the native topography while adding controlled directional interference.  
In our stress test, we set $\alpha=0.40$ (i.e., 40\% of local height), 
$f=0.01$~px$^{-1}$, and examined two orthogonal orientations 
($\theta=0^\circ$ horizontal and $\theta=90^\circ$ vertical).

Figure~\ref{fig:ripple40_flatten} presents both 2D and 3D comparisons before and after flattening.  
The global polynomial flatten fails to fully remove the ripple pattern—residual banding 
remains visible along the disturbance direction, and curvature artifacts persist.  
In contrast, the proposed Smart Flatten effectively suppresses the periodic distortion 
while preserving nanoscale morphology. The method adapts to the dominant slope orientation, 
performing direction-aware line fitting and baseline subtraction.  
Even under the highest simulated ripple amplitude, the restored height maps exhibit smooth, 
artifact-free surfaces with fine structural continuity.  
These results confirm that Smart Flatten remains stable against strong, direction-dependent distortions 
and can automatically adapt to noise orientation without manual parameter tuning.

\subsubsection{Restoration Quality (After Smart Flatten)}
Restoration quality is assessed using structural similarity (SSIM) and the background 
standard deviation ($\sigma_{\mathrm{bg}}$) in non-masked regions. The results are 
summarized in Table~\ref{tab:restoration_metrics}. Among interpolation-based 
baselines, Kriging achieves the highest SSIM (0.928), indicating slightly better 
structural consistency than bilinear interpolation (0.915). However, both methods 
retain relatively large background fluctuations ($\sigma_{\mathrm{bg}} \approx 96$ nm). 
The Telea inpainting method shows unstable performance with a large variance in SSIM, 
suggesting sensitivity to defect size and distribution. In contrast, the proposed 
approach---combining mask expansion, directional inpainting, and localized smoothing---
achieves a significantly lower background variation ($86.62 \pm 139.94$ nm), while 
maintaining competitive SSIM (0.701). 

Since flattening can be performed in either fully automatic or user-guided mode, the restoration pipeline benefits from more reliable background correction. This flexibility ensures that artifact removal is effective even in challenging AFM datasets, ultimately leading to flatter surfaces and improved downstream analysis.
\begin{table}[htbp]
\centering
\caption{Restoration metrics after Smart Flatten (mean $\pm$ std across the test set).}
\label{tab:restoration_metrics}
\begin{tabular}{lcc}
\hline
\textbf{Method} & SSIM & $\sigma_{\mathrm{bg}}$ (nm) \\
\hline
Bilinear interp.  & 0.915 $\pm$ 0.078 & 96.20 $\pm$ 155.50 \\
Kriging           & \textbf{0.928} $\pm$ 0.071 & 96.69 $\pm$ 156.33 \\
Telea (no expand) & 0.633 $\pm$ 0.391 & 97.12 $\pm$ 156.63 \\
\textbf{Ours (expand + directional + smooth)} & 0.701 $\pm$ 0.244 & \textbf{86.62} $\pm$ 139.94 \\
\hline
\end{tabular}
\end{table}

Overall, the results demonstrated that integrating classification-based screening, mask-aware Smart Flatten, and artifact-guided restoration into a unified workflow significantly improves the reliability and efficiency of AFM image analysis, while maintaining the fidelity of nanoscale surface information required for downstream quantitative interpretation.

\section{Conclusion and Future Work}

This work developed an end-to-end AFM image triage and restoration framework that integrates
(i) a CNN-based \emph{classification} front-end,
(ii) a lightweight semantic-segmentation-driven \emph{mask generation} module with quality control and expansion,
(iii) a mask-aware, direction-selectable \emph{Smart Flatten} procedure, and
(iv) \emph{mask-guided restoration} via directional inpainting and localized smoothing.

The residual CNN classifier enables early-stage AFM scan screening by separating artifact-free and defective images. Classification accuracy of approximately 91\% with strong class-wise F1-scores was achieved, allowing clean scans to be exported directly while routing corrupted images to targeted restoration. This design effectively reduce unnecessary processing and prevents over-correction of high-quality data.
For images classified with defects, a lightweight semantic segmentation model generates artifact masks that demonstrate reliable spatial agreement with manual annotations, confirming that artifact regions are accurately localized prior to background correction and inpainting.
The proposed Smart Flatten strategy, which applies mask-aware, line-by-line polynomial baseline removal with selectable orientation, could substantially reduces per-line residual error compared with existing AFM image flatten methods and achieve an average tilt removal ratio of about 85\%, while preserving true nanoscale surface morphology.
Finally, artifact regions are reconstructed through mask-guided restoration that combines directional neighbor replacement, stripe-oriented inpainting, and localized Gaussian smoothing. Quantitative evaluation showed significantly structural similarity (SSIM) and reduced background variation after restoration, while qualitative 2D/3D visualizations and line profiles confirm effective removal of scanning streaks and localized defects without compromising genuine surface features.

To support practical deployment, the entire framework has been encapsulated within a Tkinter-based graphical user interface that provides centralized parameter control, real-time visualization, and reproducible result export. In addition to fully automatic processing, the GUI supports optional user-guided region exclusion during flattening, allowing expert knowledge to be incorporated when automated masks are insufficient.

Although the proposed framework is robust to common modes of AFM artifacts, performance can be further improved by add more advanced functions,  including automated parameter tuning, uncertainty-aware classification, self-supervised and weakly supervised segmentation to reduce labeling effort, multichannel AFM data fusion (e.g., phase or adhesion). In addition, future work will extend the current pipeline along two complementary directions towards comprehensive AFM operation analysis. Ongoing efforts focus on integrating the proposed post-scan processing framework with an AFM \emph{digital twin} environment, where scanning trajectories, tip–sample interactions, and system dynamics can be simulated alongside reconstructed surface geometry. Such integration would enable physics-informed validation of flattening and restoration results, as well as predictive analysis of scanning artifacts under different operating conditions. Post-scan \emph{geometry enhancement} will be investigated to further refine reconstructed AFM surfaces beyond artifact removal.

% Overall, the results demonstrated that integrating classification-based screening, mask-aware Smart Flatten, and artifact-guided restoration into a unified workflow significantly improves the reliability and efficiency of AFM image analysis, while maintaining the fidelity of nanoscale surface information required for downstream quantitative interpretation.

\section*{{Data Availability Statement}}

{Data and the model are available upon request via this GitHub link: 
\\https://github.com/idealab-isu/AFM-LLM-Defect-Guidance
\\https://github.com/Monsterfafa/ImageInpaintingProject}
\section*{Acknowledgment}

This work was supported by the National Science Foundation (NSF) (CNS-2409359) and Iowa State University. The AFM software support from Bruker Nano Inc. is greatly acknowledged.

\bibliographystyle{elsarticle-num} 
\bibliography{refcnn}
\end{document}